\documentclass[accepted]{uai2021} % for initial submission
\usepackage{nicefrac}       % compact symbols for 1/2, etc.
\usepackage{amssymb,amsmath}
\usepackage{amsthm,amsfonts}      % blackboard math symbols
\usepackage{bbold} %For \mathbb{1}

\theoremstyle{plain}
\newtheorem{lemma}{Lemma}

\newtheorem{definition}{Definition}

\theoremstyle{remark}

\newtheorem{theorem}{\bf{Theorem}}
\makeatletter
\newtheorem*{rep@theorem}{\rep@title}
\newcommand{\newreptheorem}[2]{%
	\newenvironment{rep#1}[1]{%
		\def\rep@title{\textbf{#2 \ref{##1}}}%
		\begin{rep@theorem}}%
		{\end{rep@theorem}}}
\makeatother
\newreptheorem{theorem}{Theorem}
\newreptheorem{lemma}{Lemma}

\usepackage{tabularx}
\usepackage{algorithm,algorithmic}

\usepackage{tikz}
\usetikzlibrary{automata}
\usetikzlibrary{topaths}
\usetikzlibrary{shapes}
\usetikzlibrary{arrows}
\usetikzlibrary{decorations.markings}
\usetikzlibrary{intersections}
\usetikzlibrary{backgrounds}

\tikzstyle{utility}=[diamond,draw=black,draw=blue!50,fill=blue!10,inner sep=0mm, minimum size=8mm]
\tikzstyle{select}=[rectangle,draw=black,draw=blue!50,fill=blue!10,inner sep=0mm, minimum size=6mm]
\tikzstyle{hidden}=[dashed,draw=black,fill=red!10]
\tikzstyle{RV}=[circle,draw=black,draw=blue!50,fill=blue!10,inner sep=0mm, minimum size=6mm]
\tikzstyle{place}=[circle,draw=black,draw=blue!50,fill=blue!20,inner sep=0mm, minimum size=9mm]

\newcommand \defn {\mathrel{\triangleq}}

\renewcommand \Pr {\mathop{\mathbb{P}}\nolimits}

\newcommand \disc {\gamma}

\newcommand \reward {\mathcal{R}}

\newcommand \horizon {H} %Replaced by T

\newcommand{\expect}{\mathbb{E}}

\usepackage{mathtools}
\DeclarePairedDelimiter\ceil{\lceil}{\rceil}
\DeclarePairedDelimiter\floor{\lfloor}{\rfloor}

\def\clap#1{\hbox to 0pt{\hss#1\hss}}

\usepackage[american]{babel}
\usepackage{natbib} % has a nice set of citation styles and commands
\usepackage{mathtools} % amsmath with fixes and additions
\usepackage{booktabs} % commands to create good-looking tables
\usepackage{tikz} % nice language for creating drawings and diagrams

\title{Adaptive Belief Discretization for POMDP Planning}

\author[1]{\href{mailto:Harry Q. Bovik <harryq@example.edu>?Subject=Your UAI 2021 paper}{Divya Grover}{}}
\author[1,2]{Christos Dimitrakakis}
\affil[1]{%
    Department of Computer Science and Engineering\\
    Chalmers University, Sweden
}
\affil[2]{%
    Department of Informatics\\
    University of Oslo, Norway
}

\begin{document}
\maketitle

\begin{abstract}
Partially Observable Markov Decision Processes
(POMDP) is a widely used model to represent the
interaction of an environment and an agent, under
state uncertainty. Since the agent does not observe
the environment state, its uncertainty is typically
represented through a probabilistic belief. While
the set of possible beliefs is infinite, making exact
planning intractable, the belief space’s complexity
(and hence planning complexity) is characterised
by its covering number. Many POMDP solvers
uniformly discretize the belief space and give the
planning error in terms of the (typically unknown)
covering number. We instead propose an adaptive
belief discretization scheme, and give its associated
planning error. We furthermore characterise the
covering number with respect to the POMDP parameters. This allows us to specify the exact memory requirements on the planner, needed to bound
the value function error. We then propose a novel,
computationally efficient solver using this scheme.
We demonstrate that our algorithm is highly competitive with the state of the art in a variety of
scenarios.
\end{abstract}

\section{Introduction}

We are interested in sequential decision making problems, where an agents is interacting with a partially observable environment, meaning that it is unaware of the actual state of the environment. This interaction can be formalised through a Partially Observable Markov Decision Process (POMDP)~\citep{smallwood1973optimal}. Since the agent does not know the true state of the environment, it must instead maintain a subjective belief representing its uncertainty about the current environment state. This is typically expressed as a probability distribution over the states.

Online POMDP algorithms plan ahead by starting from the current belief and building a tree-like structure in the space of possible future beliefs, by enumerating different environment responses to the agent's actions. However, exact solutions to some lookahead horizon require an exponential size tree with respect to that horizon. Typical approximations include action selection heuristics that only consider promising  actions and leaf node approximations, that truncate the tree and replace the value of the leaf nodes with heuristics, or upper or lower bounds. While both these techniques are effective when building shallow trees, we instead propose an online POMDP algorithm that builds deep trees. In particular, we propose an adaptive discretization scheme that generates a cover with a small size, thus reducing memory and computational complexity.

\subsection{Related work}
Current state-of-the-art online POMDP solvers include POMCP~\citep{silver2010monte} and DESPOT~\cite{ye2017despot}. Like many older~\citep{ross2007bayes} and newer~\citep{kurniawati2016online} works they are based on inference using particle filtering. In building the lookahead tree, they keep its size in check by only considering ‘promising’ sequence of actions, by maintaining a statistical estimate of the action values (known as Upper Confidence Bound). DESPOT in addition, uses importance sampling to not miss important beliefs. But neither algorithm gives a structured way to check the approximate inference's effect on the planning error. While the belief discretization we propose results in a form of approximate inference as well, the focus here is on how the approximations made in the inference and in the planning process \emph{interact}.

A different method for approximate inference in POMDPs includes  belief representation methods like ~\citep{roy2005finding}. These implicitly define a kernel function over the belief space, but give no bound on the resultant solver's planning error. 

One concept we use in our theoretical analysis is the covering number of the belief space. It is a measure of the density of a set of points in any continuous space. Only recently has it been used as a complexity measure for the POMDP. In particular,~\citet{zhang2012covering} performs an experimental estimation of the covering number for various environments, by using breadth first search (BFS). However, unlike this paper, the authors do not provide a POMDP solver or derive any dependence between the underlying POMDP parameters and the covering number of the belief space.

Covering number was also used in the analysis of SARSOP~\citep{kurniawati2008sarsop} by ~\cite{lee2007makes}. In particular, it shows how the planner's error depends on the (unknown) covering number. Such solvers still depends exponentially on the lookahead, as noted by~\citep{zhang2012covering} (refer Lemma 1). Finally,~\citet{zhang2014covering} also propose PGVI, which like SARSOP, uses depth first search (DFS) to estimate the optimal action. Analytically, they are the same, since PGVI's complexity is identical to that of SARSOP.\footnote{Verify that the tuned parameters are set to identical values in both works.}

\paragraph{Contributions:} 
\begin{enumerate}
    \item We propose explicit lookahead dependent covers for the belief space. This significantly reduces the memory requirement on the planner. 
    \item We give the exact dependence of the planning error on this adaptive discretization scheme.
    \item We give an upper bound on the memory requirement of the planner by giving bounds on the covering number of the belief space.
    \item We propose a novel, computationally efficient planner with direct control on the planning error with respect to its tuning parameters. 
\end{enumerate}

\paragraph{Organization.} In Section 2, we define the POMDP model, state the used equations and introduce the concept of cover sets. In Section 3, we follow up with analysis, giving a proof of sketch first, then stating our main result, and finally the proofs. In Section 4, we describe our proposed algorithm and Section 5 is dedicated to experiments.

\section{Setting}
We divide the section into three parts: First giving the definitions of POMDP model, second giving the equations used to compute various quantities of interest in a POMDP and third part introducing the concept of $\epsilon$-cover.
\subsection{Model Definition}
\begin{definition}[MDP]
A Markov Decision Process (MDP) is a discrete-time stochastic process that provides a formal framework for decision making agents. An MDP $\mu \defn (S,A,T,R,\disc)$ is composed of a state space $S$, an action space $A$, a reward distribution $R$ and a transition function $T$. The transition function $T \defn \Pr(s_{t+1}|s_t ,a_t)$ dictates the distribution over next states $s_{t+1}$ given the present state-action pair $(s_t, a_t)$. The reward distribution $R \defn \Pr(r_{t+1}|s_t ,a_t)$  dictates the obtained reward with $r \in [0,\reward_{\max}]$.
\end{definition}
The discount $\disc \in [0,1)$ signifies the effective horizon $\horizon \defn \frac{1}{1 - \disc}$ of the agent's interaction with its environment. \emph{Henceforth, we consider only finite state, action and observation spaces}.
\newline

In many interesting settings, the agent can only indirectly estimate the state $s_t$ via a set of observations $o_t\in O$ and their (state-dependent) emission probability. Formally, 
\begin{definition}[POMDP]
A Partially Observable MDP, or POMDP, denoted by $(S,A,T,R,,\disc,O,Z)$ tuple is an MDP model augmented with an observation $o_t\in O$ set and their corresponding emission probabilities $Z\defn\Pr(o_t|s_t)$. 
\end{definition}
In such cases, the best an agent could hope to achieve is to act optimally in an expected sense (averaging over its estimation of the current state). Formally, the objective is to find action maximizing 
\begin{equation}
 V^*(b_0) \defn \lim_{\horizon \rightarrow \infty} V_{\horizon}(b_0)
\end{equation}
where $V_{\horizon}(b_0)$, called the belief value function, is defined as 
\begin{align}
	% \[
	V_{\horizon}(b_0)
	=
	\max_a \expect ( \sum_{k=0}^{H-1} \disc^{k} r_{k+1}|b_0,a) \label{eqn:val}
	% \]
	\\
	b_0 \in \Delta^{|S|} \notag \\
	\Delta^{|S|} \defn \{ b \in {[0,1]}^{|S|} \; | \; \sum_s b(s) = 1 \}
\end{align}
Where $b_0$ denotes its current state estimation, known as belief. Formally, belief is a probability distribution over the state space whose element $b(s)$ gives the probability that the system’s true state is $s$. It is a compact representation of the agent's complete history of interaction with its environment, i.e, the action-observation sequence it followed.
\newline

\begin{figure*}[th!]
		\centering
			\begin{tikzpicture}
			\node[RV] at (-5,0) (bt) {$b_t$};
			\node[RV] at (5,0) (bt1) {$b_{t+1}$};
			\node[select] at (-5,-1) (ah) {$a$};
			\node[select] at (5,-1) (ah1) {$a$};
			\draw[->] (bt) to (ah);
			\draw[->] (bt1) to (ah1);
            %First Layer - left tree			
			\node[RV] at (-4,-2) (bo1) {$b_{o_2}$};
			\node[RV] at (-6,-2) (bo2) {$b_{o_1}$};
			\draw[->] (ah) to (bo1);
			\draw[->] (ah) to (bo2);
			%First Layer - Right tree
			\node[RV] at (6,-2) (bo11) {$b_{o_2}$};
			\node[RV] at (4,-2) (bo12) {$b_{o_1}$};
			\draw[->] (ah1) to (bo11);
			\draw[->] (ah1) to (bo12);
			%Second Layer - left tree
			\node[select] at (-4,-3) (ahp1) {$a$};
			\node[select] at (-6,-3) (ahp2) {$a$};
			\draw[->] (bo1) to (ahp1);
			\draw[->] (bo2) to (ahp2);
			\node[RV] at (-5,-4) (bo3) {$b_{o_4}$};
			\node[RV] at (-7,-4) (bo4) {$b_{o_3}$};
			\draw[->] (ahp1) to (bo3);
			\draw[->] (ahp2) to (bo4);
			\draw[dashed,->] (ahp1) to (-3,-4);
			\draw[dashed,->] (ahp2) to (-5.5,-4);
			%Second Layer - Right tree
			\node[select] at (6,-3) (ahp11) {$a$};
			\node[select] at (4,-3) (ahp12) {$a$};
			\draw[->] (bo11) to (ahp11);
			\draw[->] (bo12) to (ahp12);
			\node[RV] at (7,-4) (bo13) {$b_{o_4}$};
			\node[RV] at (3,-4) (bo14) {$b_{o_3}$};
			\draw[->] (ahp11) to (bo13);
			\draw[->] (ahp12) to (bo14);
			\draw[dashed,->] (ahp11) to (5.5,-4);
			\draw[dashed,->] (ahp12) to (4.5,-4);
			%Mid drawing
			\draw[thick,->] (bt) -- (bt1) node[midway,above] {$t$};
			\node[RV] at (0,-2) (h1) {$d = 1$};
			\draw[->] (0,0) to (h1);
			\draw[dashed,-] (h1) to (-2,-2);
			\draw[dashed,-] (h1) to (2,-2);
			
			\node[RV] at (0,-4) (h2) {$d = 2$};
			\draw[->] (h1) to (h2);
			\draw[dashed,-] (h2) to (-2,-4);
			\draw[dashed,-] (h2) to (2,-4);
			\end{tikzpicture}
			\caption{Online POMDP planning}\label{fig:tree}
\end{figure*}
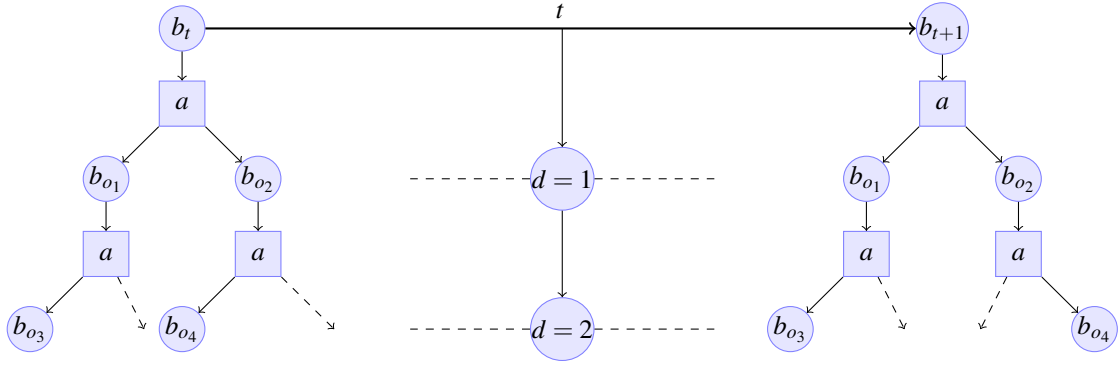

Figure~\ref{fig:tree} shows the interaction of an agent with its environment. At each time step, it evaluates the best possible action by anticipating the effect of a sequence of actions on future belief and reward. This process of anticipation is called planning and the corresponding tree-like structure is called the belief tree. Although in Figure~\ref{fig:tree}, we only calculate effect of a fixed action, you can still note the exponential growth of the number of nodes with depth. 
\subsection{Equations}
Here we describe the \textbf{standard} computational equations of the POMDP model.
To compute a new belief $b_{k+1}$, given the current belief $b_k$, action $a$ and next observation $o_{k+1}$, one uses the following set of equations:
\begin{align}
   \Pr(s_{k+1}) = \sum_{s'} T(s_{k+1}|s',a) b_k(s') \label{eq:marginal}\\
    b_{k+1}(s_{k+1}) = \frac{1}{\eta} \; Z(o_{k+1}|s_{k+1}) \Pr(s_{k+1}) \label{eq:posterior}\\
    \eta \defn \sum_{s_{k+1}} Z(o_{k+1}|s_{k+1}) \Pr(s_{k+1}) \notag
\end{align}
collectively referred to as the \emph{Bayes filter}. Equation~\ref{eq:marginal} is state marginal distribution after transition has occurred under the given action, while equation~\ref{eq:posterior} gives the Bayesian posterior (called the next belief) of the states.
\newline
Solving a POMDP involves first generating the belief tree, then doing backward induction~\citep{bellman1952theory} on it to calculate the optimal action at the root node. The corresponding Bellman equation for POMDP is:
\begin{equation}\label{eq:backward-induction}
 V(b_k)
	=
	\max_a \expect_{\Pr(r_{k+1},o_{k+1}|a,b_k)} ( \; r_{k+1} + V(b_{k+1}) \; )
\end{equation}
Where
\begin{align*}
    \Pr(r_{k+1},o_{k+1}|a,b_k) &= \Pr(o_{k+1}|a,b_k) \Pr(r_{k+1}|a,b_k) \\
    \Pr(r_{k+1}|a,b_k) &= \sum_{s'} R(r_{k+1}|s',a) b_k(s') \\
    \Pr(o_{k+1}|a,b_k) &= \eta \\
\end{align*}
Note that $b_{k+1}$ is uniquely determined by $o_{k+1}$ using Bayes filter.

\subsection{Cover sets and covering number}
For a set $\mathcal{X} \subset \mathbb{R}^n$, define metric $d_p: x \times y \rightarrow \mathbb{R} \; | \; x,y\in \mathcal{X}$:
$$ d_p(x,y) \defn {\left( \sum^n_{i=1} |x_i - y_i|^p \right)}^{\frac{1}{p}} $$
\begin{definition}
Given a metric space $(\mathcal{X},d_p)$. Define $\epsilon$-ball of a point $x_0 \in \mathcal{X}$ as $$c(x_0,\epsilon) \defn \{ x \in \mathcal{X} \; | \; d_p(x_0,x) < \epsilon \}$$
Then for a given set $B \subset \mathcal{X}$, define an $\epsilon$-cover of $B$ as 
%$$\hat{C}_B(\epsilon) \defn \{x \in \mathcal{X} \; | \; \forall b \in B, \exists \; x \in \hat{C}_B(\epsilon), d_p(x,b) < \epsilon \}$$
$$C_B(\epsilon) \defn \{x \in B \; | \; B \subset \cup \; c(x,\epsilon) \}$$
Similarly an improper cover of $B$ is defined as
\begin{align*}
    \hat{C}_B(\epsilon) \defn \{x \in \mathcal{X} \; | \; B \subset \cup \; c(x,\epsilon) \}
\end{align*}
Therefore the elements of an improper cover do not necessarily have to come from set $B$ itself.\\

The covering number is defined as $$\mathcal{N}(\epsilon,B,d_p) \defn \min  |C_B(\epsilon)| $$ i.e, the minimum cardinality sets amongst such sets.\\

An improper covering number is similarly defined
$$\hat{\mathcal{N}}(\epsilon,B,d_p) \defn \min  |\hat{C}_B(\epsilon)|$$
A standard result used here is:
\begin{equation}
\mathcal{N}(2\epsilon,B,d_p) \leq \hat{\mathcal{N}}(\epsilon,B,d_p) \leq \mathcal{N}(\epsilon,B,d_p) \label{eq:improperCover}    
\end{equation}
\end{definition}

Note that we refer to an $\epsilon$-cover as a cover set when the value $\epsilon$ is either implied or not important in the context.

\section{Analysis}
Our analysis relies on creating a sequence of covers for the belief, so that covers further away in the future are sparser. We show that we can uniformly bound the value function error by an appropriate increase in the sparsity that depends on the depth and discount factor.

Our main theorem, the \emph{planning error theorem}, gives a bound on the number of representative beliefs we need to keep track of, to bound the value function error at the root. This is accompanied by a lemma bounding the memory required by our discretization scheme and algorithm.

In particular, for exact $\horizon$-step lookahead planning, the corresponding belief tree would have order $O({|A|}^\horizon{|O|}^\horizon)$ nodes. This is prohibitive to store in memory even for a moderate horizon. A remedy to this problem comes from the fact that belief values functions are Lipschitz continuous.
While previous work~\citep[c.f.][]{lee2007makes} used this property to obtain a uniform cover, in our work we make the cover sparser as we go deeper in the tree.
%\begin{lemma}[Lipschitz Continuity]\label{prop1}
%For any two belief nodes $b$ and $b'$, we have
%$$ {||V_\horizon (b) - V_\horizon (b')||}_1 \leq L \; {||b-b'||}_1$$
%where $L = \frac{\reward_{\max} }{1-\disc}$.
%\end{lemma}

The Lipschitz property allows us to control tree size by expanding only a certain representative node's subtree and substituting its value for all beliefs represented by that particular belief node. We can then relate the approximation errors at different depths through discounting~\citep[c.f.][]{kearns2002sparse}.
Combined together these ideas gives the required planning error theorem.
%($\epsilon_{d} \leq \disc \epsilon_{d+1}$) can be used to . 

%We improve upon previous results \cdcomment{CD: Why is this not in related work?} in three ways:
%\begin{enumerate}
%    \item We propose depth dependent $\epsilon_d$-covers for each depth and give our solver's total planning error.
%    \item We improve Lemma~\ref{prop1} in Lemma~\ref{lemma:continuity} by giving depth-dependent Lipschitz constant $L$.\cdcomment{A proposition in a lemma? What do you mean? How do you improve it?}
 %   \item We give an explicit bound on the total memory required.
%\end{enumerate}
\subsection{Main Result}
\begin{definition}\label{def:discretization}
A $\epsilon$-discretization of a belief space $\Delta^{|S|}$ is defined\footnote{Ignoring the normalization constant.} as:
$$A(\epsilon) = \{ \sum_{i=1}^{S} \alpha_i {\mathbb{1}}_i | \; \alpha_i \in \{0,\epsilon,2\epsilon,\ldots,1 - \epsilon,1\}  \}$$
i.e, a vector space with ${\mathbb{1}}_i$ as basis and coefficients $\alpha_i$. ${\mathbb{1}}_i$ denotes a unit vector with ${\mathbb{1}}_i(j) = 0 \; | j\neq i$. 
\end{definition}
The error between root belief's value function to its true value function is given by the planning error theorem:
\begin{theorem}\label{theorem-1}
For any belief $b_0$, and target error $\epsilon_v$, there exists a belief tree $\mathcal{T}$ of height $h_0$ rooted at $b_0$, with separate $\epsilon_d$-discretization of the belief spaces at each depth $d$, given by the sequence
$\epsilon_d \defn \frac{\epsilon_{b_0}}{\disc^d}$, such that $b_0$'s value $\hat{V}(b_0)$ at the root satisfies
$$|\hat{V}(b_0) - V^*(b_o)| \leq \epsilon_v$$
for parameter values:
\begin{align*}
h_0 = \log_{\disc} \frac{ {(1-\disc)} \epsilon_v}{2 \reward_{\max}} \\
\epsilon_{b_0} = \frac{ {(1-\disc)} \epsilon_v}{2 \disc \reward_{\max} h_0}
\end{align*}
\end{theorem}
Contrast $\epsilon_{b_0}$ with the value of parameter $\delta (= \frac{ {{(1-\disc)}^2} \epsilon_v}{2 \disc \reward_{\max} } )$ in ~\citep{lee2007makes}. Both denote the level of discretization required. While our scheme demands a finer initial discretization ($\epsilon_{b_0}$) at the root (by a factor of $\frac{1}{h_0 (1-\disc)}$), it quickly (because $\epsilon_d \defn \frac{\epsilon_{b_0}}{\disc^h}$) becomes coarser (better) as we expand the tree deeper.
\newline
Now we state our Lemma giving total memory required to build the discretized belief tree $\mathcal{T}$.
\begin{lemma}\label{lemma:memory}
For a tree $\mathcal{T}$ of total height $h_0$ with separate $\epsilon_d$- discretized belief spaces at each level $d$, as defined in Theorem~\ref{theorem-1}. The total memory required $M_{h_0}$ is bounded by:
$$M_{h_0} \leq {\ceil{\frac{2}{\epsilon_{b_0} (1-\disc)}}}^S$$
\end{lemma}
Note that the memory requirement of SARSOP and PGVI is the covering number of the minimal $\delta$-cover for their specified tolerance value $\delta$~\citep{lee2007makes}. But they don't give an upper bound on its value, because the corresponding cover sets are only implicit.
\subsection{Proofs}
Since most proofs depend on backward induction equation~\ref{eq:backward-induction}, we use the following indexing convention: For a belief tree of total height $h_0$, we define its height by index $i$, hence $i=0$ denotes the leaf nodes level, while $i=h_0$ denotes the root level. Also for depth $d$, the relation $d = h_0 - i$ holds. This convention is also followed in the previous literature.
\newline
We first give a Lemma on Lipschitz continuity of value function at different heights:
\begin{lemma}[Lipschitz continuity]\label{lemma:continuity}
For any two belief nodes $b$ and $b'$ at height $i$, we have
$$ {||V_i (b) - V_i (b')||}_1 \leq L_i \; \delta_i$$
where 
\begin{align*}
L_i \leq \mathcal{R}_{\max} \frac{1-\disc^i}{1-\disc} + \disc^i L_0 \\
\delta_i = {||b-b'||}_1
\end{align*}
\end{lemma}
Proof is a simple consequence of H$\ddot{o}$lder's inequality.
\begin{align*}
    {||V_i (b) - V_i (b')||}_1 &\defn \\
    &= \sum_s V_i(s) (b(s)-b'(s)) \\
    &\leq {||V_i(s)||}_{\infty} \times {||b(s)-b'(s)||}_1 \\
    &= \max_s V_i(s) \times \delta_i \\
    &= L_i \delta_i
\end{align*}
where second line follows from H$\ddot{o}$lder's inequality.
\newline
Hence by using equation~\ref{eq:backward-induction} we get
\begin{align*}
L_i &\defn \max_s V_i \\
\max_s V_i &= \max_s \; ( r + \disc V_{i-1} ) \\
\max_s V_i &=  \mathcal{R}_{\max} + \disc \max_s V_{i-1}
\end{align*}
which applied recursively gives the result $\square$
\newline

We next give a Lemma relating value function error across different heights:
\begin{lemma}\label{lemma:error}
Maximum error for any node at height $h$, denoted by $\epsilon_h$ is given by:
$$ \epsilon_h \leq \sum^{h-1}_{i=0} \disc^{h-i} L_i \delta_i + \disc^h \epsilon_0$$
where $\epsilon_0$ denotes the error at leaf nodes.
\end{lemma}
Proof. For a belief $b$ at level $i=0$, consider its error in value function $\epsilon(b)$ as:
\begin{align*}
\epsilon(b) &= |\hat{V}(b) - V^*(b)| \\
&\leq |\hat{V}(b) - V(b')| + |V(b') - V^*(b)| \\
&\leq L_0 \delta_0 + \epsilon_0
\end{align*}
Where line 2 comes from the fact that it maybe the case that value of $b$ is substituted by value of another belief $b'$, constituting the $\delta_0$-cover at level $i=0$. Moreover, for any parent of $b$ (say $b_i$), at level $i=1$, its error $\epsilon_1 \leq \disc \epsilon(b)$~\citep{kearns2002sparse}. Hence by recursion, we get:
\begin{align*}
    \epsilon_1 &\leq \disc (L_0 \delta_0 + \epsilon_0) \\
    \epsilon_2 &\leq \disc (L_1\delta_1 + \disc L_0 \delta_0 + \disc \epsilon_0) \\
    &\ldots \\
    \epsilon_h &\leq \sum^{h-1}_{i=0} \disc^{h-i} L_i \delta_i + \disc^h \epsilon_0    
\end{align*}
Where $\epsilon_i$ denotes the error in value function for any belief at level $i$.
\paragraph{Proof of Theorem~\ref{theorem-1}:} Equating root error in Lemma~\ref{lemma:error} to target error $\epsilon_v$ gives:
$$ \sum^{h-1}_{i=0} \disc^{h-i} L_i \delta_i + \disc^h \epsilon_0 = \epsilon_v $$
Combining this with Lemma~\ref{lemma:continuity} gives:
$$ \sum^{h-1}_{i=0} \disc^{h-i} \left( \reward_{\max} \frac{1-\disc^i}{1-\disc} + \disc^i L_0 \right) \delta_i + \disc^h \epsilon_0  = \epsilon_v$$
Assuming $L_0 = \frac{\reward_{\max}}{1-\disc}$, and leaf error $\epsilon_0 = \frac{\reward_{\max}}{1-\disc}$, we get
\begin{align*}
\sum^{h-1}_{i=0} \disc^{h-i} \epsilon_0 (1-\disc^i + \disc^i) \delta_i + \gamma^h \epsilon_0 &= \epsilon_v\\
\disc^h \epsilon_0 \left( \sum^{h-1}_{i=0} \frac{\delta_i}{\disc^i} + 1 \right) &= \epsilon_v \\
\left( \sum^{h-1}_{i=0} \frac{\epsilon_{b_0}}{\disc^{h-i-1} \disc^i} + 1\right)&= \frac{\epsilon_v}{\disc^h \epsilon_0}\\
\frac{h \epsilon_{b_0}}{\disc^{h-1}} + 1 = \frac{\epsilon_v}{\disc^h \epsilon_0}
\end{align*}
Where we use the fact that 
\begin{align*}
\delta_i &\defn \epsilon_d = \frac{\epsilon_{b_0}}{\disc^d} = \frac{\epsilon_{b_0}}{\disc^{h-i-1}}
\end{align*}
Finally, substituting 
$$h=h_0=\log_{\disc} \frac{\epsilon_v}{2\epsilon_0} = \log_{\disc} \frac{\epsilon_v (1-\disc)}{2 \reward_{\max}}$$
gives $\epsilon_{b_0} = \frac{\disc^{h_0-1}}{h_0} \; \square$
\newline

We now calculate the total memory required to store a discretized tree $\mathcal{T}$ by proving Lemma~\ref{lemma:memory}. Note that $A(\epsilon)$ acts as $\epsilon$-cover for the set ${[0,1]}^{|S|}$. Since belief space $\Delta^{|S|} \subset {[0,1]}^{|S|}$, $A(\epsilon)$ also forms an improper cover for it. We bound the covering number for each depth $d$, by bounding the cardinality of the corresponding sets $A(\epsilon_d)$. We consider the $\epsilon_d \defn \frac{\epsilon_b}{\disc^d}$-discretization of any arbitrary sequence of ${[0,1]}^{|S|}$ spaces.
\paragraph{Proof of Lemma~\ref{lemma:memory}:} For any $h_0$ length sequence of discretized spaces, total memory required:
\begin{align*}
\hat{M}_{h_0} &\defn \sum_d \mathcal{N}(\epsilon_d,A(\epsilon_d),d_1) \\
&= \sum_d |A(\frac{\epsilon_b}{\disc^d})| \\
log \sum_d |A(\frac{\epsilon_b}{\disc^d})| &\leq |S| \sum_d \log(1 + \floor{\frac{\disc^d}{\epsilon_b}}) \\
&\leq |S| \sum_{d=0}^{h_0} \floor{\frac{\disc^d}{\epsilon_b}} \:\: | \; \disc^{h_0} \geq \epsilon_b \\
&\leq |S| \sum_{d=0}^{h_0} \frac{\disc^d}{\epsilon_b} \\
\hat{M}_{h_0} &\leq {\ceil{\frac{1}{\epsilon_b (1-\disc)}}}^{|S|}
\end{align*}
From equation~\ref{eq:improperCover}
$$\mathcal{N}(\epsilon,\Delta^{|S|},d_1) \leq \mathcal{N}(\epsilon/2,{[0,1]}^{|S|},d_1)$$
Combing the two results proves the Lemma $\square$
%\newline
%In practise, we can set $m_h \propto 1/\epsilon_h$.

%\subsection{Comparison to existing literature}

%Past literature have proposed approximate planners with order squared dependence on $\cnumber$, but they don't characterize $\cnumber$'s explicit dependence on POMDP parameters. 
%\newline
%discretizing the belief space, thereby controlling the memory budget allocated. The planner approximates the true posterior while

\section{Finite Memory Planner}
We now introduce our main algorithm, Finite Memory Planner (FMP). This is an approximate online POMDP algorithm, with the following two main steps:
\begin{enumerate}
    \item Building an approximation $\mathcal{T}$ to the full belief tree, achieved here by recursively calling Algorithm~\ref{algo1}.
    \item Calculating the optimal action at the root and its value function using equation~\ref{eq:backward-induction-modified} on $\mathcal{T}$.
\end{enumerate}
The lookahead algorithm for building the belief tree $\mathcal{T}$ is described in Algorithm~\ref{algo1}. 

\begin{algorithm}[t!]
		\caption{FMP (Finite Memory Planner) \\ {\bfseries Parameters:} Horizon $h_0$, discretization levels $\epsilon_d$}
		\begin{algorithmic}[1]
			\STATE {\bfseries Input:} Nodes with possible beliefs $b_d\in B_d$ at depth $d$,\\ tolerances $\epsilon_d$, current height $h$
			\IF{$h=h_0$}
			\STATE return []
			\ENDIF
			\STATE children = []
			\FOR{all $b_d \in B_d$}
			\FOR{all actions a}
			\FOR{all next observations $o_{d+1}$}
			\STATE $b_{d+1},o_{prob},r =$ BayesFilter$(b_{d},o_{d+1},a)$
			\label{inference-step}
			\STATE $\hat{b}_{d+1} = NN(b_{d+1},A(\epsilon_{d+1}))$\label{nn-step}
			\STATE children.append$([\hat{b}_{d+1},o_{prob},a,r])$
			\ENDFOR
			\ENDFOR
			\ENDFOR
			\STATE FMP(children,$\epsilon_{d+1}$,$h+1$)
		\end{algorithmic}
		\label{algo1}
	\end{algorithm}

\paragraph{Building the tree.}
The first stage is building a tree to a fixed depth, as follows.
\begin{enumerate}
    \item For each belief $b_d$ at depth $d$, action $a$ and observation $o$, we call ‘BayesFilter’ function at line~\ref{inference-step}, which uses equations~\ref{eq:posterior} and \ref{eq:marginal} to compute its true posterior and other attributes. They are later needed to compute the root value and optimal action using backward induction (equation~\ref{eq:backward-induction-modified}).
    \item At line~\ref{nn-step}, we replace the true posterior $b_{d+1}$ by its nearest neighbour in the $A(\epsilon_d)$ set which serves as its $\epsilon_d$-cover. This is computationally cheap because of uniform discretization (refer Definition~\ref{def:discretization}).
\end{enumerate}
We remark that the computational cost of Bayesian inference (line~\ref{inference-step}) is a shared burden for all POMDP solvers, and is not any more expensive for FMP, than for other solvers. One possible advantage of FMP is that inference can be quite flexible, e.g., one may use exact, variational inference or sampling based inference at line~\ref{inference-step}.\footnote{Contrast this to POMCP, which is solely based on Monte Carlo (particle filter) inference.}
Finally, we make a note that even $\epsilon$-cover based algorithms SARSOP and PGVI are significantly different. In contrast to FMP, both perform DFS to build the tree, thereby forgoing any chance of building a minimal $\epsilon$-cover at each depth. Computationally, FMP is cheaper as it does not need to cluster points at any level, while both SARSOP and PGVI do.
\paragraph{Calculating the value.} 
In our scheme, the subsequent beliefs $b_{k+1}$ of any belief $b_k$ are replaced by some approximations $\hat{b}_{k+1}$ in the $\epsilon_{k+1}$-cover. Hence, the backwards induction equation is now
\begin{equation}\label{eq:backward-induction-modified}
 V(b_k)
	=
	\max_a \expect_{\Pr(r_{k+1},o_{k+1}|a,b_k)} ( \; r_{k+1} + V(\hat{b}_{k+1}) \; )
\end{equation}
When the observation space is large we approximate the above by sampling $o_{k+1} \sim \Pr(o_{k+1}|a,b_k)$ and $r_{k+1} \sim \Pr(r_{k+1}|a,b_k)$.\footnote{In particular,  we sample $(o_{k+1},r_{k+1})$ in \emph{RandomPOMDP} environment but only $(r_{k+1})$ in \emph{RockSample} environment since it has binary observation space.}
Note that the algorithm returns the complete approximate value function for every depth $k+1$ to the previous depth $k$.

\section{Experiment}
We compare with State-of-the-art planner POMCP~\citep{silver2010monte} mainly, on RandomPOMDP and RockSample environments. We also compare with PBVI~\citep{pineau2003point} on RandomPOMDP and DESPOT~\citep{ye2017despot} on RockSample environment. The environment description is as follows:
\begin{enumerate}
    \item \textbf{RandomPOMDP ($RP[s,sp]$):} A random POMDP of given state, action and observation space. The reward, transition and observation distributions is generated using a sparsity parameter $sp \in \{0.3,0.6,0.9\}$ with $\mathcal{R}_{\max} = 1$. We experiment with different state space sizes $|S|$ and sparsity parameters. We fix $|Z|=|S|$ and $|A|=4$. Denote a particular configuration by $RP[s,sp]$.
    \item \textbf{RockSample ($RS[n,k]$):} Environment proposed in ~\citep{smith2012heuristic}, its state space size ranges from 12 thousand to 7 million. Although the belief space is very sparse, consisting of only 8 to 15 partially observed quantities. RS[n,k] denotes a map of size $n \times n$, searched by a agent, looking to collect good rocks between k possible total rocks. Its has a binary observation space, with distance dependent observation likelihood, giving more accurate observations as the agent moves closer to the rocks. Its goal is to collect the good rocks and exit the map. The agent knows its location and that of rocks, but is unaware of the state (good/bad) of each rock.
\end{enumerate}
\subsection{setup}
We take discount $\disc=0.95$ (standard) and run each experiment for $90$ steps\footnote{Inline with POMCP and DESPOT paper for RockSample, and sufficient for RandomPOMDP (since $0.95^{90} \approx 0.01$)}. We perform 50 experiments for each (Planner,POMDP) combination. We use POMDPy library~\citep{emami2015pomdpy} for RockSample simulator and PyPOMDP library~\citep{PyPOMDP} to solve our generated RandomPOMDP environment with POMCP and PBVI. While for the RockSample environment, we simply report the published (and tuned) results for DESPOT and POMCP (Table.1 in  ~\citep{ye2017despot}). Like \citep{smith2012point}, we mask the known states in RockSample environment. Finally, we also use GNU parallel~\citep{tange2018gnu} to run parallel experiments.
\paragraph{Parameter tuning:} It is worth noting that there are minimal tuning parameters to begin with, just $h_0$ and the sequences $\{\epsilon_d\}$. The solution quality is fairly linear with fineness of discretization, and inversely proportional to the error tolerance levels. In the experiments, the values are selected such that computation remains within satisfactory online planning limit (5 second for all environments up to RS[11,11], while 30 seconds for RS[15,15]). In particular, the following values were selected: $\epsilon_d = \{\epsilon_0 = \frac{1}{5},\ldots,\epsilon_H = \frac{1}{5}\}$ or $\{\epsilon_0 = \frac{1}{5},\epsilon_1 = \frac{1}{4},\ldots,\epsilon_H = \frac{1}{2}\}$ (depending on planning limit) and tree depth $H=5$ or $H=4$ (depending on planning limit). Note that the $\epsilon_d$-sequence and $H$ chosen are arbitrary and gave sufficiently good results without finetuning. For RandomPOMDP, we the allocate maximum possible budget (simulation time) to POMCP\footnote{We tuned for exploration constant $c\in \{3,10,100\}$.} and PBVI.
\newline

The source code is available online and with supplementary material. The general FMP planner is made available in a pseudo-code style implementation in Python language.

\subsection{Results and discussion}
\begin{table}[!th]
    \centering
    \begin{tabular}{|c|c|c|c|}
 \hline
 RS[n,k] & RS[7,8] & RS[11,11] & RS[15,15]\\
 \hline
 |S| & 12,544 & 247,808 & 7,372,800 \\
 \hline
 |A| & 13 & 16 & 20 \\
 \hline
 FMP & 20.24$\pm$0.85 & 21.49$\pm$0.63 & 15.67$\pm$0.37 \\
 \hline
 POMCP$^*$ & 20.71$\pm$0.21 & 20.01$\pm$0.23 & 15.32$\pm$0.28 \\ 
 \hline
 DESPOT$^*$ & 20.93$\pm$0.30 & 21.74$\pm$0.30 & 18.64$\pm$0.28 \\
 \hline
    \end{tabular}
    \caption{Total Discounted Reward. $^*$Reported values.}
    \label{table:rocksample}
\end{table}

\begin{table}[!ht]
    \centering
    \begin{tabular}{|c|c|c|c|}
    \hline
    RP[s,sp] & FMP & POMCP & PBVI \\
         \hline
    RP[100,0.9] & 6.02 & 5.22 & 5.82 \\
         \hline
    RP[100,0.6] & 4.38 & 3.99 & 4.65 \\
         \hline
    RP[100,0.3] & 2.42 & 1.57 & 2.40 \\
         \hline
    RP[60,0.9] & 6.04 & 5.90 & 5.01 \\
         \hline
    RP[60,0.6] & 4.65 & 4.94 & 4.23 \\
         \hline
    RP[60,0.3] & 2.42 & 2.67 & 2.27 \\
         \hline
    RP[30,0.9] & 6.36 & 6.37 & 6.26 \\
         \hline
    RP[30,0.6] & 4.12 & 5.33 & 4.26 \\
         \hline
    RP[30,0.3] & 3.40 & 3.53 & 3.74 \\
         \hline
    \end{tabular}
    \caption{Total Discounted Reward.}
    \label{table:RandomPOMDP}
\end{table}

The results are given in Table~\ref{table:rocksample}  and \ref{table:RandomPOMDP} for RockSample and RandomPOMDP\footnote{For RandomPOMDP, the variance was low and we do not report it simply for clarity.} respectively. It is clear that the performance of FMP is competitive in all settings, even with relatively lax parameters. For RockSample, FMP's python implementation is significantly handicapped compared to reported values of POMCP and DESPOT, obtained by parameter tuning and with much faster C++ implementation. We believe FMP further would benefit from such speed optimization. The RandomPOMDP performance gives some hints to FMP behaviour. It is evident that FMP performs best when there are many beliefs to begin with, i.e, it performs best in low sparsity setting. For high sparsity, it performed competitively to the rest (for larger state sizes), and we guess it misses out in other cases simple due to less lookahead (only $H=5$, again due to planning limit and Python language implementation).
\newline

Finally, we contrast FMP with other planners by noting their tuning and implementation complexity. For example, POMCP requires depth of tree, number of particles used for inference, simulation budget, UCB constant, as parameters. DESPOT in addition to all that, requires maintaining upper/lower bounds on belief nodes. Most other solvers apply such heuristics as well.
\section{Conclusion}

We have a presented a simple algorithm for online planning in POMDPs that is highly competitive with the state-of-the-art despite its apparent simplicity. The main idea behind FMP was to design an algorithm where the inference and planning are tightly coupled to minimise the estimation error. In particular, the algorithm performs approximate inference with adaptive belief discretization so as to upper bound the planning error, with the coarseness of the approximation carefully tuned so as not to overly increase the size of the planning tree. This makes the algorithm both theoretically attractive and practical.
\newline

In future work, there is possibility to extend in multiple directions. we would like to extend our method to continuous spaces, which can be done as the proof strategy only needs kernel function over beliefs. For example, kernels for the sufficient statistics of the belief over continuous state space could be explored. The analysis could also be extended to get regret bounds in Reinforcement Learning setting. Both directions may lead to potential solvers that can handle a wide variety of problems.

\bibliography{uai2021-template}

\appendix

\end{document}